\documentclass{article}
\PassOptionsToPackage{round}{natbib}

\usepackage[section]{placeins}

\usepackage[preprint]{neurips_2021}
\usepackage[utf8]{inputenc} 
\usepackage[T1]{fontenc}    
\usepackage{hyperref}       
\usepackage{url}            
\usepackage{booktabs}       
\usepackage{amsfonts}       
\usepackage{nicefrac}       
\usepackage{microtype}      
\usepackage{xcolor}         
\usepackage{amsmath}
\usepackage{amssymb}
\usepackage{mathdesign}

\usepackage{algorithm}
\usepackage{algorithmic}
\algsetup{linenosize=\tiny}

\usepackage{graphicx}
\usepackage{subfig}
\usepackage{array, booktabs, makecell}

\DeclareMathOperator*{\argmin}{argmin}

\usepackage{natbib}
\title{Adaptive Cost-Sensitive Learning in Neural Networks for Misclassification Cost Problems}

\author{%
  Ohad Volk$^*$ \\
  Faculty of Engineering\\
  Bar-Ilan University\\
  \texttt{volkoha@biu.ac.il} \\
   \And
   Gonen Singer$^*$ \\
   Faculty of Engineering\\
  Bar-Ilan University\\
   \texttt{gonen.singer@biu.ac.il} \\
}

\begin{document}

\maketitle
\begin{abstract}
We design a new adaptive learning algorithm for misclassification cost problems that attempt to reduce the cost of misclassified instances derived from the consequences of various errors. Our algorithm (adaptive cost sensitive learning – AdaCSL) adaptively adjusts the loss function such that the classifier bridges the difference between the class distributions between subgroups of samples in the training and test data sets with similar predicted probabilities (i.e., local training-test class distribution mismatch). We provide some theoretical performance guarantees on the proposed algorithm and present empirical evidence that a deep neural network used with the proposed AdaCSL algorithm yields better cost results on several binary classification data sets that have class-imbalanced and class-balanced distributions compared to other alternative approaches.
\end{abstract}

\section{Introduction}

Cost-sensitive learning is a domain that takes the misclassification costs resulting from various errors into account when training machine learning models. Research studies in this domain have designed sensitive classifiers such that insensitive models are modified to be biased toward classes with high misclassification costs. These types of machine learning models aim to achieve minimum misclassification costs \cite{domingos1999metacost, elkan2001foundations,ling2004decision}. Usually, these approaches, which directly incorporate cost-sensitive capabilities into a learning algorithm, are specific to a group of classification methods and not generalizable.

Several research studies used a cost-sensitive learning approach to adjust insensitive deep neural networks by modifying misclassification costs based on the training data set,  so that they can handle imbalanced data \cite{ raj2016towards,wang2016training, sze2017weight, buda2018systematic,zhang2018cost,dong2020cost}. These studies usually used the misclassification costs to reflect the imbalance of the classes in the data, i.e., the ratio between misclassification costs reflects the ratio between classes in the training data set. 

In many real-life problems, there are many situations in which both class imbalance and misclassification costs exist. For example, when classifying the severity of disease, the ratio between the cost of incorrectly classifying a severely ill patient as a healthy one and the cost of classifying a healthy patient as a severely ill one can be much higher than the ratio between healthy and severely ill patients in the data set. This paper addresses the above issues and proposes a cost-sensitive learning method for classification applications with misclassification error costs and class imbalances.
The cost-learning parameters used to handle misclassification costs and class imbalance are considered differently in the learning process. In order to handle the class imbalance, the misclassification costs are used to modify the deep neural network toward the minority classes, and other learning parameters are used to reduce the misclassification costs. In addition, whereas all aforementioned studies, usually modified the misclassification costs based on the training data set to adjust deep neural networks, we propose a different approach, in which the loss function is adjusted adaptively, such that the classifier would be directed to reduce local high misclassification costs in the validation data set. In this setup, we develop an adaptive cost-sensitive learning (AdaCSL) algorithm that optimizes the loss function between consecutive learning epochs. 

This paper makes four main contributions:
\begin{enumerate}
\item We propose a cost-sensitive learning method for classification applications with misclassification error costs and class-imbalanced problems. 
\item  We propose an adaptive algorithm that applies the cost-sensitive learning method, addressing local high misclassification costs in the validation data set 
by adjusting adaptively the loss function. The proposed algorithm can guide the design of a variety of different methods and loss functions. 
\item The proposed algorithm applied in a convolutional neural network (CNN) and had been shown to outperform baseline and other cost-sensitive methods (over a test data set), based on four data sets reflecting different class imbalance ratios.
\end{enumerate}
The remainder of this paper is organized as follows. In Section 2, we briefly discuss the related work on cost-sensitive learning approaches and the required adaptation to our problem. In Section 3, we introduce our proposed cost-sensitive learning method for handling local high misclassification costs in the validation data and describe the adaptive cost-sensitive learning (AdaCSL) algorithm. Section 4 presents our experiments and results when evaluating and benchmarking the proposed AdaCSL with a CNN against baseline and other cost-sensitive methods. Finally, Section 5 concludes this paper.

\section{Preliminaries}

We now review the cost-sensitive binary classification problem for individual instances in a training data set. Then we generalize the cost-sensitive approach for subgroups of samples with the same positive probability, aiming to identify local high misclassification costs. 

\subsection{The Cost-Sensitive Approach}
First, we introduce notations used throughout the paper. Consider binary classification in which examples are taken from the instance space denoted by $X \in \mathcal{X} \subseteq \mathcal{R}^d$, where $d$ represents the number of features, and the output is denoted by $Y \in \{0,~1\}$, where 1 is the positive class and 0 is the negative class. A classification model $\mathcal{F}: \mathcal{X} \rightarrow [0,~1]$ maps an instance with input feature $X$ to a positive probability $P(Y=1|X)$, or, briefly, $P(1|X)$, between $0$ and $1$, such that a higher probability expresses a stronger belief that an instance relates to the positive class. The negative probability, $P(Y=0|X)$, or briefly, $P(0|X)$, expresses the belief that an instance relates to a negative class, and also ranges between $0$ and $1$. Given a threshold value $T$ and a positive probability $P(1|X)$, the instance is classified to the positive class if $P(1|X) > T$, and to the negative class otherwise.

In practical classification applications, inevitably, misclassification error costs are introduced. These costs derive from the consequences of misclassification errors.
Cost-sensitive research takes these costs into account using three main approaches:
\begin{enumerate}
\item Data level approaches: Changing the distribution of classes, usually as pretraining approaches, using over- or under-sampling within the training data set \cite{elkan2001foundations, ling2008cost}, or weighting  the training examples according to the misclassified samples, as post-training boosting approaches \cite{masnadi2010cost}. 
\item Algorithmic level approaches: Directly incorporating cost-sensitive capabilities into existing insensitive learning methods to be biased toward classes with high misclassification costs 
\cite{ling2008cost,lin2017focal,wu2018weighted,aurelio2019learning}. 
\item Decision level approaches: Using the original training data set and insensitive models, while adjusting the decision threshold $T$ after modeling a classifier to minimize the misclassification costs \cite{hernandez2013roc, buda2018systematic}.
\end{enumerate}

Assume $\mathcal{C}$ represents a misclassification cost matrix. The  $\mathcal(i,j)$ entry represents the misclassification cost of predicting instance $X$ as class $i$ when the actual class is $j$, denoted by $\mathcal{C}_(i,j)$. 
The costs are constant across all instances in the data set for each type of classification error.
Following \cite{ling2008cost}, who proved that any cost matrix could be converted into a matrix without incurring a cost to correct classifications such that the decision of classifying an example as positive will not be changed, we assume that $\mathcal{C}_(i,~j)=0$, if $i=j$. The optimal prediction for an instance $X$ is the class $i$ that minimizes the total expected cost (risk), given by $\mathcal{R}\left(X,~i\right) = \sum_j P(j|X) \mathcal{C}_{(i,~j)},$  
where $P(j|X)$ is the probability that an instance $X$ relates to class label $j$. Thus. the optimal prediction that minimizes the cost will be defined by $i = \argmin\limits_i \mathcal{R}(X,~i)$. It is shown \cite{elkan2001foundations} that an optimal threshold (a decision level approach), which yields the minimum expected cost for a single instance, is defined by,
$
T = \frac{\mathcal{C}_{(1, ~0)}}{\mathcal{C}_{(0,~1)} + \mathcal{C}_{(1,~0)}}.
$
Insensitive learning algorithms try to achieve maximum accuracy while classifying instances according to the class with the majority probability and according to a fixed threshold of 0.5 for binary problems. Thus, it is proved \cite{elkan2001foundations} that the number of negative class samples in the training data set should be multiplied by $\left(\frac{T}{1-T}\right)\left(\frac{1-T'}{T'}\right)$ to make a threshold $T$ corresponds to another threshold $T'$  (a data level approach), while  $T'=0.5$ generates a well-balanced data set. 

Several research studies developed algorithmic level methods to convert specific insensitive classification models into sensitive ones, by incorporating cost-sensitive capabilities into the learning algorithm, thus attempting to reduce the total cost of misclassified instances. Most of these studies focus on decision trees models \cite{ting2002instance,maimon2014data, zhao2017cost, chaabane2019enhancing}. In the context of deep neural networks, few papers tackle the data imbalance problem inherent in the learning process, while explicitly focusing on specific loss functions \cite{wang2016training, khan2017cost, geng2018cost, wu2018weighted,aurelio2019learning}. None of them, however, consider misclassification cost problems or problems with both misclassification costs and class imbalance. Furthermore,  all existing studies, modified the deep neural networks based only on the training data, without considering misclassification costs from the validation data. 

Our cost-sensitive learning hybrid method, in contrast, combines an algorithm-level approach and a decision-level approach. The algorithm level approach modifies the learners' loss function to alleviate their bias toward majority groups (negative class samples), by using the misclassification costs,  while the decision level approach adaptively adjusts the loss function such that the classifier would be directed to reduce local high misclassification costs in the validation data set. The proposed method can be applied to any classifier with a loss function and is not restricted only to neural networks. 

In Section 2.2, we divide a data set into subgroups of samples with the same predicted probabilities. We then prove that the threshold $T$ holds for all instances in a training data set. This proof and the local subgroup concept constitute the basis for deriving the proposed adaptive cost-sensitive learning method for modifying deep neural networks so they address the performance degradation caused by local high misclassification costs. 

\subsection{Cost-Sensitive Approach For Subgroups Containing Samples With The Same Predicted Probabilities}
Let $\mathcal{D}^t = (x_k,~y_k), k=1,...,K$ be a training data set of labeled instances. Assume $\mathcal{D}_n^t, n=1,...,N$ presents a subgroup $n$ of samples from the data set that are mapped to the same positive probability $P(1|x_l) = P^t_n(1), \forall x_l \in \mathcal{D}_n^t $ by a classification model.

\textbf{Theorem 1.} \textit{The optimal threshold for classifying instances from a data set is constant for all subgroups $\mathcal{D}_n^t, n=1,...,N$ and is equal to}

\begin{equation}
T = \frac{\mathcal{C}_{(1,~0)}}{\mathcal{C}_{(1,~0)} + \mathcal{C}_{(0,~1)}}
\end{equation}

\textit{Proof.} This proof follows the threshold proof for individual instances from \cite{elkan2001foundations}. Let $|\mathcal{D}_n^t|$ represent the number of instances in a subgroup. Assume an optimal threshold $T_n^t$ for classifying samples in subgroup $\mathcal{D}_n^t$, which yields the minimum misclassification costs. 
Thus, if $P^t_n(1) < T_n^t$, all instances belonging to subgroup $\mathcal{D}_n^t$ will be classified as negative class samples, and the cost of the errors emanating from the subgroup will be $\sum_{x_l \in \mathcal{D}_n^t} \mathcal{R}(x_l,i=0) = -P_n^t(1) \cdot |\mathcal{D}_n^t| \cdot \mathcal{C}_{(0,~1)}$. If $P_n^t(1) > T_n^t$,  all instances belonging to subgroup $\mathcal{D}_n^t$ will be classified as positive class samples, and the cost will be  $\sum_{x_l \in \mathcal{D}_n^t} \mathcal{R}(x_l,i=1) = -(1-P_n^t(1)) \cdot |\mathcal{D}_n^t| \cdot \mathcal{C}_{(1,~0)}$. The optimal threshold for making optimal decisions will be attained by comparing the two expressions and is equal to $T_n^t = \frac{\mathcal{C}_{(1,~0)}}{\mathcal{C}_{(0,~1)} + \mathcal{C}_{(1,~0)}}$. Since the optimal threshold does not depend on subset $\mathcal{D}_n^t$ or the positive probability $P_n^t(1)$, the optimal threshold for all instances in the training data set is equal, i.e., $T=\frac{\mathcal{C}_{(1,~0)}}{\mathcal{C}_{(1,~0)}+ \mathcal{C}_{(0,~1)}}.\hfill\ensuremath{\blacksquare}$


Insensitive learning algorithms try to achieve maximum accuracy in the applied phase by minimizing the loss in the training phase. After the training phase, to evaluate the accuracy of the model's prediction, the samples are classified according to the class with the greater probability and according to a fixed threshold of 0.5 for binary problems. In contrast, the decision-level approach for cost-sensitive problems, which seeks to decrease misclassification costs, classifies samples according to different values of a threshold calculated using the problem's misclassification costs, as proved in Theorem 1. Thus, to achieve a specific and constant threshold  $T'$ which does not necessarily depend on the costs (a specific case, for example, can be $T'=0.5$), we follow \cite{elkan2001foundations} to find the expression in which we should multiply the number of negative samples in each subgroup. In Lemma 1, we prove that this expression is equal for all subgroups.

\textbf{Lemma 1.}\textit{ When the optimal thresholds of subgroups in the training data set are $T_n^t=T, \forall n=1,...,N$, to make a target probability threshold of a training data set correspond to a given probability threshold $T$', the number of negative examples in every subgroup $\mathcal{D}^t_n$ in the training data set should be multiplied by, }
\begin{equation}
\left(\frac{\mathcal{C}_{(1,~0)}}{\mathcal{C}_{(0,~1)}}\right)\left(\frac{1-T'}{T'}\right)    
\end{equation}

\textit{Proof idea.} This lemma is a generalization of Theorem 1 in \cite{elkan2001foundations} for subgroups of samples, and follow his proof. The main idea is that similar to Elkan, the number of negative class samples in a subgroup $\mathcal{D}^t_n$, should be multiplied by $\left(\frac{T_n^t}{1-T_n^t}\right)\left(\frac{1-T'}{T'}\right)$ to make a threshold $T_n^t$ correspond to another threshold $T'$. Assuming the number of negative instances in each subgroup is 
 $\sum_{l=1}^{|\mathcal{D}_n^t|}\left(1-y_l\right)$, the new number of negative samples in the training data set will be $\sum_{n=1}^{N}\left(\frac{T_n^t}{1-T_n^t}\right)\left(\frac{1-T'}{T'}\right)\sum_{l=1}^{|\mathcal{D}_n^t|}\left(1-y_l\right)$. Following Theorem 1, with constant $T_n^t$ and assuming the same given threshold for all subgroups $T'$, we obtain  
$\left(\frac{T_n^t}{1-T_n^t}\right)\left(\frac{1-T'}{T'}\right) \sum_{n=1}^{N}\sum_{l=1}^{|\mathcal{D}_n^t|}\left(1-y_l\right) $, where $\sum_{n=1}^{N}\sum_{l=1}^{|\mathcal{D}_n^t|}\left(1-y_l\right) $ represents the number of negative class samples in the entire training data set. Using $T_n^t=T$ from Theorem 1, the number of negative class samples in every subgroup, or equivalently in the training data set, should be multiplied by $\left(\frac{\mathcal{C}_{(1,~0)}}{\mathcal{C}_{(0,~1)}}\right)\left(\frac{1-T'}{T'}\right).\hfill\ensuremath{\blacksquare}$
    

\section{An Adaptive Cost-Sensitive Learning Method }

This section introduces a cost-sensitive learning method in neural networks using a modification of a loss function by utilizing both a decision threshold approach and an algorithmic level approach. The first expression, found in Section 3.1, modifies the loss function to alleviate their bias towards majority groups caused by the class imbalance problem, by using the misclassification costs. The second expression, found in Section 3.2, adjusts the loss function by an adaptive cost-sensitive learning (AdaCSL) algorithm presented in Section 3.3, aiming to address local high misclassification costs in the validation data set. 
We adopt the terminology and findings presented in Section 2 and apply them to subgroups of samples in the validation data set. 

\subsection{Cost-Sensitive Loss Function }
A general cost-sensitive learning method is developed by modifying the loss function used in neural networks, aiming to reduce misclassification error costs. The proposed method is suitable for class-balanced and class-imbalanced problems with predefined misclassification costs. For this purpose, we formulate the mean loss over the training set considering the cost matrix $\mathcal{C}$ as
\begin{equation}
E(\theta) = \frac{1}{K}\sum\limits_{k=1}^{K}\mathcal{L}(y_k, \hat{y}^{\theta}_k,\mathcal{C})
\end{equation}
where the predicted output of the $k^{th}$ sample $\hat{y}^\theta_k$ is parameterized by the weights and biases $\theta$ of the network, which is denoted by $\hat{y}_k$. Different from other research studies, here the cost matrix is constant and predefined. The loss function $\mathcal{L}(\cdot)$ can be any suitable surrogate loss. 
The objective of the learning process is to find the optimal $\theta^*$, which yields the minimum possible error, expressed by $\theta^*=\argmin\limits_\theta E(\theta)$. We discuss the Cross-Entropy loss (CE) for binary classification, which measures the distance from the actual class ($0$ or $1$) to the predicted value, $\hat{y}_k = P(1|x_k)$, and can be expressed as  $\mathcal{L}(y_k, \hat{y}_k) = -y_k \log(\hat{y}_k)-(1-y_k)\log(1-\hat{y}_k)$.

\textbf{Theorem 2.}\textit{ When the optimal thresholds of subgroups of a training data set are equal $T_n^t=T, \forall n = 1,...,N$, to make a target probability threshold $T$, calculated in eq. (1), corresponding to a given probability threshold $T$', the CE loss should be}

\begin{equation}
\mathcal{L}(y_k, \hat{y}_k,\mathcal{C}) =
    -y_k \log (\hat{y}_k)-\left(\frac{\mathcal{C}_{(1,~0)}}{\mathcal{C}_{(0,~1)}}\right)\\
    \left(\frac{1-T'}{T'}\right)(1-y_k)\log{(1-\hat{y}_k)}
\end{equation}
\textit{Proof.} The total CE loss of a subgroup $\mathcal{D}_n^t$, with the same positive class probability $\hat{y}_l = \hat{y}_n^t, \forall{x_l} \in \mathcal{D}_n^t$  is 
$
\sum\limits_{l=1}^{|\mathcal{D}_n^t|}\mathcal{L} (y_l,~\hat{y}_n^t ) = -\sum\limits_{l=1}^{|\mathcal{D}_n^t|}(y_l \log (\hat{y}_n^t)+(1-y_l) \log (1-\hat{y}_n^t)).$ Since $\hat{y}_n^t$ is constant and does not depend on a sample $x_l$, we obtain,
$\sum\limits_{l=1}^{|\mathcal{D}_n^t|}\mathcal{L} (y_l,~\hat{y}_n^t )=-\log(\hat{y}_n^t) \sum\limits_{l=1}^{|\mathcal{D}_n^t|} y_l -\log (1-\hat{y}_n^t)\sum\limits_{l=1}^{|\mathcal{D}_n^t|}(1-y_l).$
The expression $\sum\limits_{l=1}^{|\mathcal{D}_n^t|}(1-y_l)$ reflects the number of negative samples in subgroup $\mathcal{D}_n^t$. The effect of changing the number of negative class samples in a subgroup as presented in Lemma 1 is equivalent to multiplying this expression in the CE loss function. Thus,  to make a probability threshold $T_n^t=T$ of a subgroup, defined in eq. (1), correspond to a given probability threshold $T'$, the loss function should be $\sum\limits_{l=1}^{|\mathcal{D}_n^t|}\mathcal{L}(y_l, \hat{y}_l)=
      -\sum\limits_{l=1}^{|\mathcal{D}_n^t|}( y_l \log (\hat{y}_l)+\left(\frac{\mathcal{C}_{(1,~0)}}{\mathcal{C}_{(0,~1)}}\right)\left(\frac{1-T'}{T'}\right)\
     (1-y_l)\log(1-\hat{y}_l)).$
Following Theorem 1, since the target threshold for all subgroups in the training data set is similar, the loss function is $\mathcal{L}(y_k, \hat{y}_k)=
      -y_k \log (\hat{y}_k)-\left(\frac{\mathcal{C}_{(1,~0)}}{\mathcal{C}_{(0,~1)}}\right)\left(\frac{1-T'}{T'}\right)\
     (1-y_k)\log(1-\hat{y}_k),\forall{(x_k,y_k)} \in \mathcal{D}^t.
\hfill\ensuremath{\blacksquare}$

\subsection{Loss Function Adjustment For Addressing Local High Misclassification Costs}
We propose to adjust the loss function between two consecutive epochs, in order to enable neural networks to reduce local high misclassification costs in the validation dataset, without changing the validation data set. 
Let $\mathcal{D}^v = (x_r,~y_r), r=1,...,R$ be a validation data set of labeled instances. Assume $\mathcal{D}_m^v, m=1,...,M$ represent subgroups of samples among the validation data set that are mapped to the same positive probability $P(1|x_j) = P_m^v(1), \forall_{X_j} \in \mathcal{D}_m^v$ by a classifier, denoted by $P_m^v$. 

\textbf{Theorem 3.}\textit{ To make an actual threshold of subgroup $T_m^{v,a}$, which yields minimum costs in the validation data set, correspond to a probability threshold $T$', the CE loss for the samples in subgroup $(x_j,~y_j) \in \mathcal{D}_m^v$  should be}

\begin{equation}
\mathcal{L}(y_j, \hat{y}_j) =
    -y_j \log (\hat{y}_j)-e^{\frac{-(T' - T^{v,a}_m)}{T'(1-T')}}\left(\frac{\mathcal{C}_{(1,~0)}}{\mathcal{C}_{(0,~1)}}\right)\left(\frac{1-T'}{T'}\right)(1-y_j)\log(1-\hat{y}_j)
\end{equation}

\textbf{Lemma 2.}\textit{ The predicted class probability $\hat{y}_k = P(1|x_k)$ of an instance $k$ should be changed to $ \hat{y}_k'$ in response to a change in the prior probability of an instance in the data set from $P  = P(1) $ to $P' = P'(1)$ such that}
\begin{equation}
\hat{y}'_k = P' \frac{\hat{y}_k - \hat{y}_k P}{P - \hat{y}_k + P' \hat{y}_k - PP'}    
\end{equation}

\textit{Proof.} Lemma 2 was proved in \cite{elkan2001foundations}, using Bayes' rule. Given this lemma, we can now prove Theorem 3.

\textit{Proof of Theorem 3.} In Theorem 2, we proved that the second expression should be multiplied by $\frac{\mathcal{C}_{(0,~1)}}{\mathcal{C}_{(1,~0)}}(\frac{1-T'}{T'})$ to yield the minimum cost at $T$' in the training data set. According to Lemma 2, since the actual probability of being positive is different from the predicted positive probability $P_m^{v,a} \neq P_m^v$, the actual threshold that yields the minimum cost is $T_{m}^{v,a}\neq T'$. Suppose that $P_m^v = \frac{1}{1 + \eta}$ and $P_m^{v,a}=\frac{1}{1 + \eta'}$; thus, the number of actual negative samples should be multiplied by
 $\frac{\eta}{\eta'} = \left(\frac{P_m^{v,a}}{1-P_m^{v,a}}\right)\Big(\frac{1-P_m^v}{P_m^v}\Big)$ to modify $P_m^{v,a}$ so that it corresponds to $P_m^v$. From Lemma 2, assuming that $ \hat{y}_{k}' = T^{v,a}_{m}$  and $\hat{y}_{k} = T'$, $P_m^{v,a}$ can be expressed as $P_m^{v,a} = \frac{T_{m}^{v,a}P_m^v\left(1-T'\right)}{T'-T'P_m^v-T'T_m^{v,a}+T^{v,a}_{m}P_m^v}$ and thus the number of actual negative samples should be multiplied by $\left(\frac{P_m^{v,a}}{1-P_m^{v,a}}\right) \Big(\frac{1-P_m^v}{P_m^v}\Big)  = \left(\frac{T^{v,a}_{m}}{1-T^{v,a}_{m}}\right)\left(\frac{1-T'}{T'}\right) $. By the first-order Taylor series expansion around the point $T'$ of the function, $\left(\frac{T^{v,a}_{m}}{1-T^{v,a}_{m}}\right)\Big(\frac{1-T'}{T'}\Big)$  yields $\left(\frac{T^{v,a}_{m}}{1-T^{v,a}_{m}}\right)\Big(\frac{1-T'}{T'}\Big) \cong e^{\frac{-\left(T'-T^{v,a}_{m}\right)}{T'\left(1-T'\right)}}$ . Thus, the total number of negative class samples $\sum\limits_{j=1}^{|\mathcal{D}_m^{v}|}(1-y_{j})$ in the loss of subgroup $\mathcal{D}_m^{v}$ should be multiplied by
\begin{equation}
    \lambda=e^{\frac{-(T'-T_m^{v,a})}{T'(1-T')}}.
\end{equation}
Similar to Theorem 2, since we use the same optimal threshold for all samples in the subgroup, each sample in the validation data set $(x_j,~y_j) \in \mathcal{D}_m^v$ should be multiplied by the same expression and the loss function should be $\mathcal{L}(y_j, \hat{y}_j,\lambda) =
       -y_j \log (\hat{y}_j)-\lambda\left(\frac{\mathcal{C}_{(1,~0)}}{\mathcal{C}_{(0,~1)}}\right)\left(\frac{1-T'}{T'}\right)(1-y_j)\log(1-\hat{y}_j).\hfill\ensuremath{\blacksquare}$

Each sample should be adjusted in the loss function using eq. (7) according to the threshold  $T_m^{v,a}$, derived from its affiliation to the relevant subgroup. To find a single averaged expression to update every sample in the validation data set, we weight the expressions in eq. (7) by the subgroup's sizes as follows:
\begin{equation}
\lambda^\psi=\frac{\sum\limits_{m=1}^{M} |\mathcal{D}_m^v| e^{\frac{-\left(T' - T_m^{v,a}\right)}{T'\left(1-T'\right)}}}{R}.    
\end{equation}
Thus, the following loss function with an averaged adjustment $\forall(x_r,y_r)\in \mathcal{D}^v$ is
\begin{equation}
    \mathcal{L}(y_r, \hat{y}_r, \lambda^\psi) =
       -y_r \log (\hat{y}_r)-\lambda^\psi\left(\frac{\mathcal{C}_{(1,~0)}}{\mathcal{C}_{(0,~1)}}\right)\\
      \left(\frac{1-T'}{T'}\right)(1-y_r)\log(1-\hat{y}_r).
\end{equation}
\subsection{Adaptive Cost-Sensitive Learning Algorithm}
Our goal is to develop an adaptive cost-sensitive learning (AdaCSL) algorithm for any neural network, aiming to achieve minimum misclassification costs. The algorithm learns the weight and bias parameters $\theta$ in the training process, while adjusting the loss function between each consecutive epoch, using the validation data set. In each epoch, we keep the loss function constant and use gradient descent with the backpropagation of eq. (3). To adjust the loss function, we use the neural networks with the current estimate of parameters $\theta$ and divide the validation data set  $\mathcal{D}^v = (x_r,~y_r), r=1,...,R$ into $M$ subgroups $\mathcal{D}_m^v, m=1,...,M$, each containing samples that are mapped to a similar positive probability $P_m^v$. For each subgroup we find the optimal thresholds $T_m^{v,a}, m=1,...,M$ among a threshold set $\mathcal{T}$ yielding the minimum misclassification costs and calculate $\lambda$ from eq. (7). In the next training epoch, the loss function is updated using $\lambda^\psi$ from eq. (8). The adaptive process ends when the loss function is not significantly adjusted, as defined by $\epsilon$.
\begin{algorithm}[tb]
        \caption{Adaptive Cost-Sensitive Learning (AdaCSL)}
\begin{algorithmic}
   \STATE {\bfseries Input:} Training data set $\mathcal{D}^t=(X^t,Y^t)$, validation data set $\mathcal{D}^v=(X^v,Y^v)$, number of batches $B$, cost matrix $\mathcal{C}$, model $\mathcal{F}$, targeted threshold $T'$, set of potential thresholds $\mathcal{T}$
   
   \STATE {\bfseries Output:} Learned parameters $\theta^*$\\
   \STATE {\bfseries Initialize:} $\theta$ arbitrarily, $i$=1, $\lambda_1^\psi=1$
   \REPEAT
   \FOR{$b=1$ {\bfseries to} $B$}  
   \STATE $\hat{Y}^t_b \leftarrow \mathcal{F}(X^t_b;~\theta)$  $~~~~~~~~~~~~~~~~~~~~~~~~~~~~~~~~~$ // \footnotesize Probability 
   \STATE $\theta \leftarrow \theta - \alpha \nabla_\theta \mathcal{L}(Y_b^t,~\hat{Y}^t_b,\lambda_i^\psi)$
   \ENDFOR
   \STATE $\hat{Y}^v$ $\leftarrow$ $\mathcal{F}\left(X^{v};~\theta\right)$ $~~~~~~~~~~~~~~~~~~~~~~~~~~~~~~~~~~~$ // \footnotesize Probability 
   \STATE
   Divide $\mathcal{D}^v$ into subgroups  $\mathcal{D}_m^v, m=1,...,M$
   \FOR{$m=1$ {\bfseries to} $M$}
   \STATE $T_m^{v,a} \leftarrow \argmin \limits_{\tau \in \mathcal{T}}~Cost\left(\hat{Y}_m^v,~Y_m^v,~\tau,~\mathcal{C}\right)$
   \ENDFOR
   \STATE 
$\lambda_{i+1}^\psi\leftarrow \lambda_{i}^\psi\cdot\frac{\sum\limits_{m=1}^{M} |\mathcal{D}_m^v| e^{\frac{-\left(T' - T_m^{v,a}\right)}{T'\left(1-T'\right)}}}{R}  $ $~~~~~~~~~~~~$ // Adaptive\\
   $~~~~~~~~~~~~~~~~~~~~~~~~~~~~~~~~~~~~~~~~~~~~~~~~~~~~~~~~~~~~~~~~~~~~~~~~$// Adjustment
   
   $i= \leftarrow i+1$
   \UNTIL $\left(\lambda_{i+1}^\psi-\lambda_{i}^\psi\right)<\epsilon$ 
\end{algorithmic}
\end{algorithm}
\medskip

\section{Experiments}
In this section, the AdaCSL algorithm used in Convolutional Neural Network (CNN) is applied in two experiments with different objectives. The first experiment aims to illustrate the algorithm’s novelty in resolving local high misclassification costs in the validation data set by adjusting the loss function between two consecutive learning epochs. The objective of the second experiment is to evaluate the AdaCSL algorithm compared to other CNNs, some with other cost-sensitive approaches. 
\subsection{Illustration Of Addressing Local High Misclassification Costs By The AdaCSL Algorithm
}
\subsubsection{Data sets, experimental setup , and method}
We illustrate the AdaCSL algorithm’s novelty in directing deep neural network to reduce local high misclassification costs in the validation data set, based on the task of distinguishing between photos of wolves and Eskimo dogs \cite{ribeiro2016should}. 
We use a class-balanced training data set with 1,200 images and class-balanced validation data set with 400 images and identical misclassification costs ${C}_{(1,~0)}={C}_{(0,~1)}=2$  (i.e., local misclassification improvement is equivalent to local accuracy improvement). We use the PyTorch \cite{paszke2019pytorch} implementation of the ResNet18 \cite{he2016deep} architecture for our experiment. Since the dataset is relatively small, we use a pre-trained version of ResNet on ImageNet \cite{deng2009imagenet} to reduce the initial feature extraction phase’s importance. We use the Stochastic Gradient Descent (SGD) algorithm to learn the parameters and regularize with weight decay. In order to illustrate the effect of the adjustment of the loss function using the AdaCSL algorithm, we use the following mechanism for two consecutive learning epochs. In Epoch 1, by using a standard Cross-Entropy (CE) as the loss function, we generated probability estimations for both training and validation data sets. We split the training samples into 10 subgroups $\mathcal{D}_m^t ,~m=1,…,10$ and the test samples into $10$ subgroups $\mathcal{D}_m^v,~m=1,...,10$, such that , $P(wolf|x) \in [0.1\cdot{m}-0.1, 0.1\cdot), \forall x \in\mathcal{D}_m = (\mathcal{D}_m^t, \mathcal{D}^v_m)$. For the following epoch 2, we build both CNN based on a standard CE (CNN w/o AdaCSL) and CNN with AdaCSL algorithm (CNN w AdaCSL), aiming to evaluate the changes in accuracy and CE loss for different subgroups. 



\subsubsection*{Results}
The accuracy (Acc), Cross-Entropy (CE) and misclassification costs (Cost) are reported in Table 1. Each sample is associated with a subgroup $m$,  following its probability estimation at epoch 1. The result of the best CNN in epoch 2, with the preferable CE and cost values for each subgroup $m$ is shown in bold. We observe that (1) the CNN with the proposed AdaCSL algorithm achieves better CE values over all validation subgroups with improvements in the range of 1.57-12.28\%, with an average of 5.98\%. (2) the CNN with the proposed AdaCSL was further improved the misclassification costs (and accuracy) in three out of four of the subgroups (subgroups $\mathcal{D}_5^{v},\mathcal{D}_7^{v},\mathcal{D}_8^{v}$) with the highest local misclassification costs in Epoch 1 (marked with an underline). In the fourth subgroup $\mathcal{D}_6^{v}$, the CNN with the AdaCSL achieves identical misclassification cost to the CNN without the AdaCSL. (3) The overall costs of the CNN without AdaCSL was 10\% higher than the CNN with the AdaCSL. These results illustrate the algorithm’s novelty in resolving local high misclassification costs in the validation data set by adjusting the loss function between two consecutive learning epochs.
\begin{table*}[htbp]
\centering
\resizebox{\textwidth}{!}{
\begin{tabular}{@{}rrrccccccccccccccc@{}}
\toprule                         &  & & 
 \multicolumn{2}{c}{$~$Epoch 1 CNN}
 &
\multicolumn{4}{c}{\hphantom{aaaaaaa} Epoch 2 CNN w/o AdaCSL} & &
\multicolumn{4}{c}{\hphantom{aaaaaaa}Epoch 2 CNN w/ AdaCSL} & 
\\

\cmidrule(lr){4-5}\cmidrule(lr){6-10}\cmidrule(lr){11-15}
 $m$ &\thead{Subgroup\\ Range} & \thead{Subgroup\\ Size} &\thead{Val Acc } & \thead{Val Cost }  & \thead{Train \\ Acc} & \thead{Val \\ Acc} &\thead{Train \\ CE} &  \thead{Val \\ CE} & \thead{Val \\ Cost} & \thead{Train \\  Acc} &  \thead{Val \\ Acc} & \thead{Train \\ CE} & \thead{Val \\ CE} & \thead{Val \\ Cost}\\
 
\midrule
2& $0.1-0.2$         & 3     & 100\% & 0    & 100\%     & 100\%     & 0.104      &0.086  & 0     & 100\%           & 100\%            & 0.093             & \textbf{0.076} & 0\\
                           
3& $0.2-0.3$         & 6     & 100\% &  0  & 98.4\%      & 100\%     & 0.168      &0.188  & 0      & 98.4\%    & 100\%            & 0.151            & \textbf{0.167} & 0\\

4& $0.3-0.4$         & 32     & 93.8\%   & 4  & 90.3\%      & 96.9\%     & 0.330      &0.279 & \textbf{2}       & 90.3\%           & 93.8\%            & 0.314             & \textbf{0.258} & 4\\

5& $0.4-0.5$         & 78     & 70.5\%  & \underline{46}  & 86.4\%      & 92.3\%     & 0.445      &0.389 & 12      & 85.9\%           & 93.6\%            &36             & \textbf{0.374} & \textbf{10}\\

6& $0.5-0.6$         & 120    & 60.0\%   & \underline{96}  & 94.1\%      & 93.3\%     & 0.458      &0.477    & 16   & 94.1\%           & 93.3\%            & 0.451             & \textbf{0.470} & 16\\

7& $0.6-0.7$         & 100   & 65.0\% & \underline{70}     & 90.5\%      & 89.0\%     & 0.460      &0.461  & 22     & 92.5\%           & 91.0\%            & 0.452             & \textbf{0.452} & \textbf{18}\\

8& $0.7-0.8$         & 55   & 67.3\%  & \underline{36}    & 81.6\%      & 76.4\%     & 0.467      &0.487 & 26      & 82.9\%           & 80.0\%            & 0.452             & \textbf{0.475} & \textbf{22}\\

9& $0.8-0.9$         & 6     & 16.7\%  & 10   & 71.4\%      & 16.7\%     & 0.471      &0.876   & 10    & 71.4\%           & 16.7\%            & 0.452             & \textbf{0.815} & 10\\
           
\bottomrule
\end{tabular}
}
\caption{A comparison of CE loss, accuracy, and cost in subgroups levels between CNN with AdaCSL and without AdaCSL.}
\label{tbl:experimentsPriorities}
\end{table*} 
\subsection{Evaluation of the AdaCSL Algorithm}
We evaluate the proposed approach on four binary-class data sets, with the imbalance ratio ranging from 1 (class-balanced problem) to 8 (class-imbalanced problem), between the majority (negative) and minority (positive) samples, as follows:

\subsubsection{Data sets, Experimental Setup , and Method}

\textbf{Melanoma Classification: Harvard Dataverse (HAM10K)} \cite{tschandl2018ham10000} This melanoma data set consists of 
1,113 melanoma positive and 8,902 melanoma negative images (1:8 ratio). 

\textbf{Breast Histopathology Images: (IDC)} \cite{janowczyk2016deep} Invasive Ductal Carcinoma (IDC) is the most common subtype of all breast cancers. This breast cancer data set consists of 78,786 IDC positive and 198,738 IDC negative images (1:2.5 ratio).

\textbf{Real and Fake Face Detection (RFFD)} \cite{RFFD} This data set comprises expert-generated high-quality 'photoshopped' face images. The data set consists of 960 positive (real) images and 960 negative (fake) images (1:1 ratio).\\
\newline
\textbf{PlantVillage-Data Set (PV-D)} \cite{mohanty2016using} This data set 
comprises images of healthy and infected leaves of crops plants.%
The data set consists of 58,299 negative (diseased) and 27,866 positive (healthy) images (1:2.1 ratio).

We evaluate the performance of the proposed algorithm  to achieve minimum costs in three setups for each data set, reflecting different ratios between misclassification costs,  $\rho = \frac{\mathcal{C}_{(0,~1)}}{\mathcal{C}_{(1,~0)}}$, such that $\rho \in\Big\{\frac{|\mathcal{D^-{}}|}{|\mathcal{D}^+|},3\left(\frac{|\mathcal{D^-{}}|}{|\mathcal{D}^+|}\right),5\left(\frac{|\mathcal{D^-{}}|}{|\mathcal{D}^+|}\right)\Big\}$, where $|\mathcal{D^+}|$ and $|\mathcal{D^-}|$ are the number of positive and negative samples in the data set, respectively.
In Table 1, for example, for the data set 'HAM10K', with a ratio between classes of $\frac{|\mathcal{D^-}|}{|\mathcal{D^+}|}=8$, we define the following three setups: $\rho = \frac{\mathcal{C}_{(0,~1)}}{\mathcal{C}_{(1,~0)}}=8/24/40$.
\begin{figure}%
    \centering
    \subfloat[\centering]{{\includegraphics[scale=0.255]{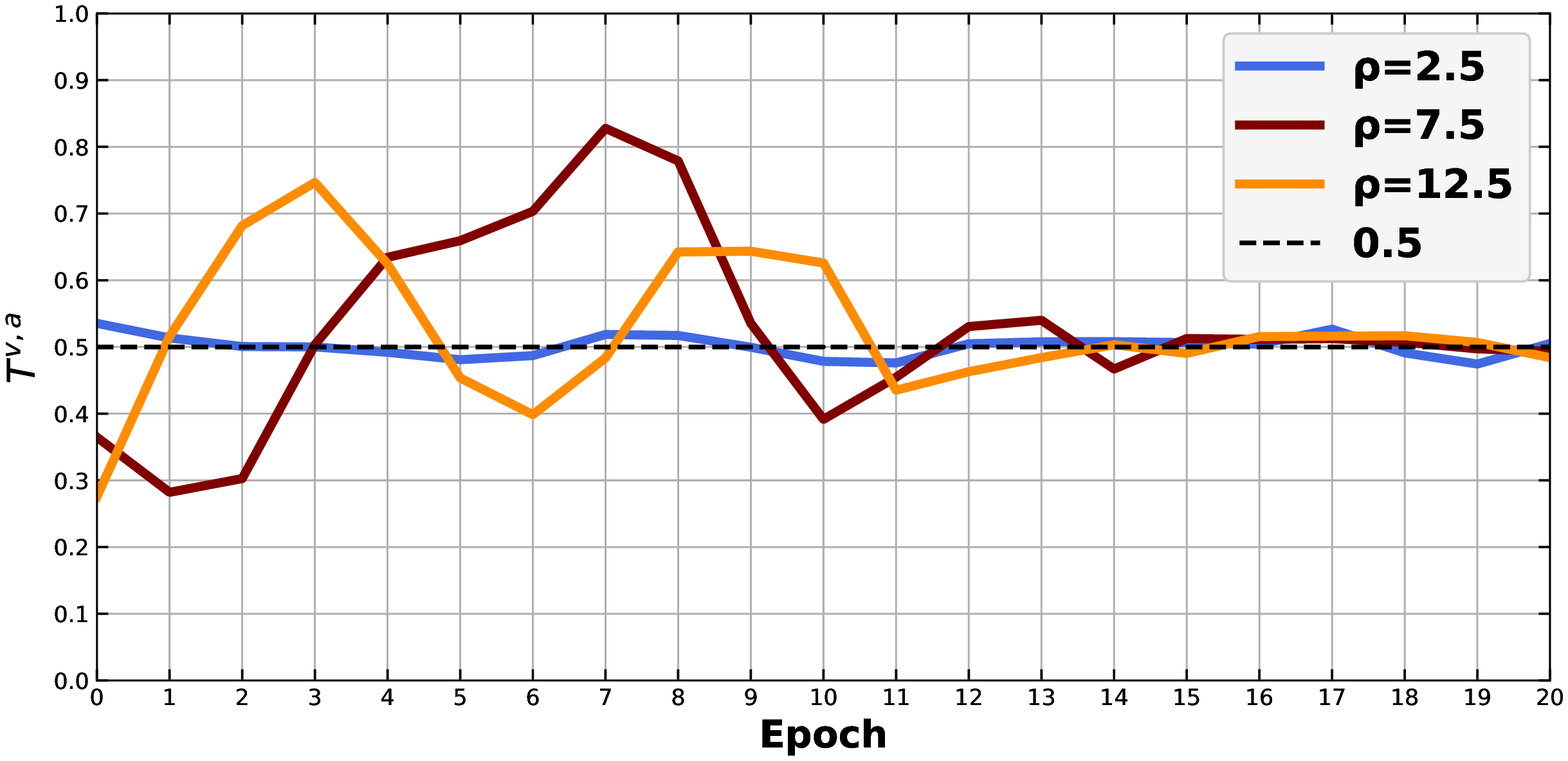} }}%
    \qquad
    \subfloat[\centering]{{\includegraphics[scale=0.26]{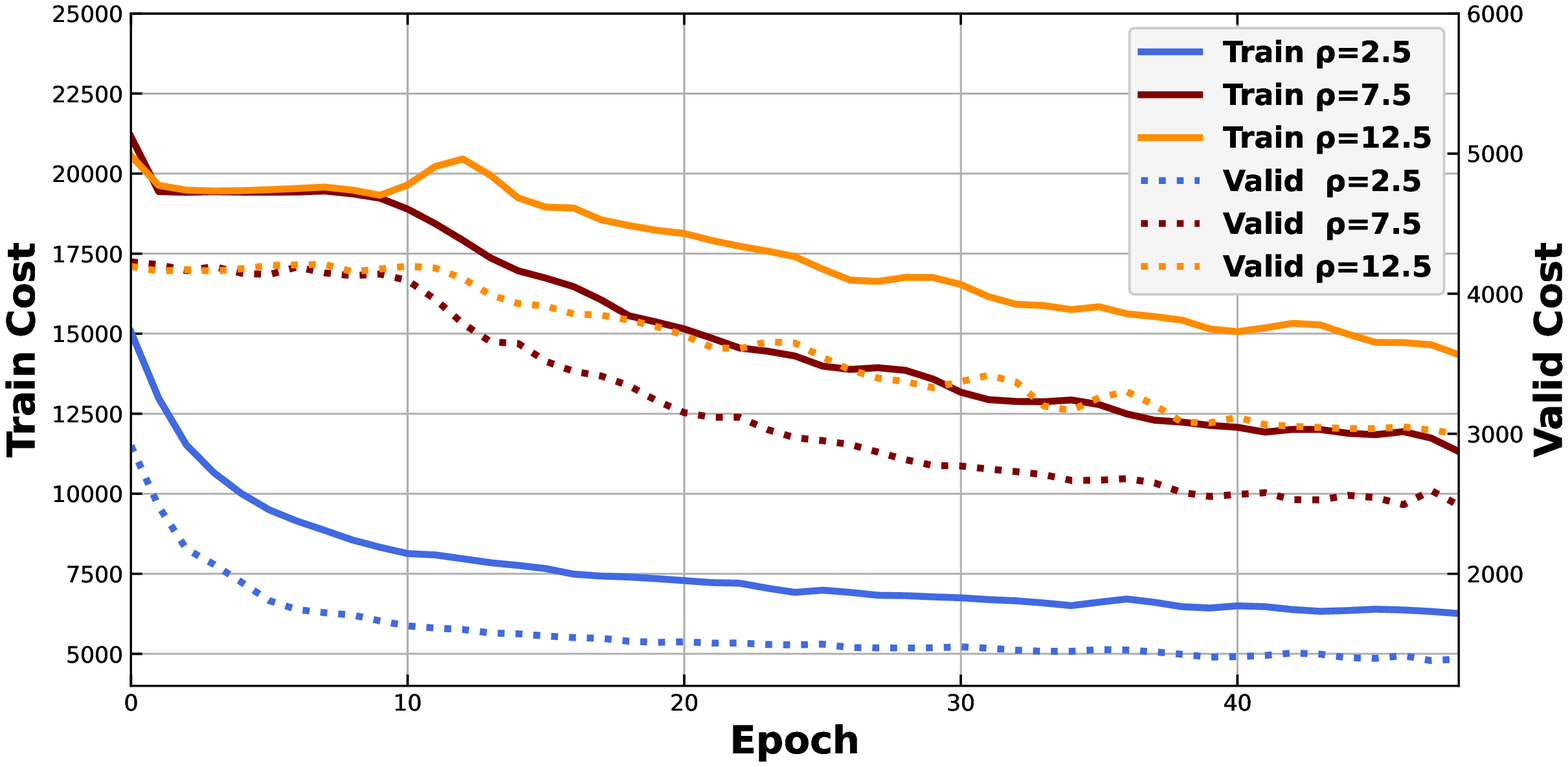} }}%
    \caption{(a): $T^{v,a}$ values oscillate around $T'=0.5$ in the IDC dataset. (b): An observed decrease in the training and validation misclassification costs on the IDC dataset for the case of the AdaCSL algorithm in different experimental setups.}%
    \label{fig:example}%
\end{figure}
We apply the proposed AdaCSL algorithm while  obtaining a single threshold $T^{v,a}$ from the entire validation data set (equivalent to assuming equal thresholds for all subgroups, $T^{v,a}_m = T^{v,a}, \forall m$). We evaluate and benchmark the results against baseline method and  other cost-sensitive approaches as following: \begin{enumerate}
    \item
      CNN with standard CE loss function as baseline method.
    \item
      CNN with threshold adjustment (CNN w/TA) as decision level approach \cite{kong2020sound,lan2020novel}.
      \item
      CNN with synthetic minority oversampling technique (CNN w/SMOTE) as a data level approach  \cite{pandey2019automatic}. 
      \item
      CNN with weighted CE (CNN w/WCE) as an algorithmic level approach \cite{wu2018weighted,aurelio2019learning}
\end{enumerate}

When experimenting with the balanced data set (RFFD), not all algorithms are applied. CNN w/ SMOTE is redundant in all the setups. Also,  CNN w/ TA and CNN w/ WCE  for the experimental setup of $\rho = 1$, are redundant since they are equivalent to the standard CNN.

\subsubsection{Results}
\begin{table*}[htbp]
\centering
\resizebox{\textwidth}{!}{
\begin{tabular}{@{}cccccccrrrrrrrrrrrrrrr@{}}
\toprule
 \textbf{Datasets}                         &  &
\multicolumn{2}{c}{CNN} & 
\multicolumn{2}{c}{CNN w/ TA} & 
\multicolumn{2}{c}{CNN w/ SMOTE} & 
\multicolumn{2}{c}{CNN w/ WCE} & 
\multicolumn{2}{c}{CNN w/AdaCSL} \\

\cmidrule(lr){3-4}\cmidrule(lr){5-6}\cmidrule(lr){7-8}\cmidrule(lr){9-10}\cmidrule(lr){11-12}
 & Experimntal Setup & Cost & Acc & Cost & Acc & Cost & Acc &  Cost & Acc &  Cost & Acc\\
 
\midrule
HAM10K&$\rho = 8$         & 1001     & 72.01\%     & 2000      & 11.83\%     & 1132      & \textbf{84.84}\%       & 824           & 74.19\%            & \textbf{658}             & 82.57\%\\
                            &  $~~\rho =24$         & 1895     & 83.25\%     & 2000        & 11.83\%     & 2198      & \textbf{85.98}\%        & 1090           & 65.75\%            & \textbf{959}              & 69.25\%\\
                            & $~~\rho = 40$            & 6012     & \textbf{88.71}\%    & 2000        & 11.83\%     & 4649     & 86.33\%        & 1400           & 57.80\%           & \textbf{1345}              & 60.25\%\\

IDC& $~~~\rho = 2.5$         & 1296     & \textbf{86.66}\%     & 1482       & 86.12\%     & 1263      & 85.85\%       & 1249           & 84.84\%            & \textbf{1225}             & 83.64\%\\
                            & $~~~\rho = 7.5$          & 2712     & 85.36\%     & 3492        & \textbf{86.00}\%     & 2557      & 83.93\%        & 1990           & 74.92\%            & \textbf{1925}              & 78.23\%\\
                            & $~~~~~\rho = 12.5$            & 4741     & \textbf{86.25}\%    & 5250        & 86.11\%     & 3910     & 84.33\%        & 2481           & 66.79\%           & \textbf{2318}              & 75.18\%\\
RFFD &$\rho = 1$         & 214     & 62.85\%     & -       & -     & -      & -       & -           & -            & \textbf{210}             & \textbf{63.54}\%\\
                            & $\rho = 3$          & 322     & 59.38\%     & 492        & \textbf{60.07\%}     & -      & -        & 282           & 51.73\%            & \textbf{274}              & 57.29\%\\
                            &$\rho = 5$            & 540     & \textbf{65.97\%}    & 538        & 62.15\%     & -     & -        & 318           & 51.73\%           & \textbf{286}              & 50.35\%\\
PV-D & $~~~\rho = 2.1$         & 406     & 98.40\%     & 12925       & 32.34\%     & 428      & 98.40\%       & 390           & 98.22\%            & \textbf{170}             & \textbf{99.50\%}\\
                            & $~~~\rho = 6.3$          & 1319     & 98.35\%     & 12925        & 32.34\%     & 1098      & 98.41\%        & 1134           & 94.26\%            & \textbf{279}              & \textbf{98.82\%}\\
                            & $~~~~~\rho = 10.5$            & 1779     & 97.27\%    & 12925       & 32.34\%     & 1224     & \textbf{98.42}\%        & 1835           & 94.12\%           & \textbf{585}              & 98.03\%\\
\bottomrule
\end{tabular}
}

\caption{A comparison of the performance of the proposed adaptive algorithm vs. other baseline and other cost sensitive methods.}
\label{tbl:experimentsPriorities}
\end{table*}
We select the model of each compared method that yields the best misclassification costs over the validation set, and report their costs and accuracy over the test set in Table 2. The results of the best model are shown in bold for each data set and experimental setup. We observe that the CNN with the proposed AdaCSL algorithm achieves the best performance over all data sets and setups with improvements in the range of 2-75\%, with an average of 20\%. Furthermore, although the proposed algorithm is designed to yield good cost results, it achieved competitive accuracy results for experimental setups that reflect an imbalance ratio equal to the cost ratio (cost-sensitive approach for imbalance problems) with the best accuracy in two of the four data sets. 
 Figure 1(a) shows the values of the actual threshold $T^{v,a}$, which yields the minimum costs in the IDC validation data set, over different epochs. It can be observed that the actual threshold converges to the desired threshold $T'=0.5$, while differences between the imbalance ratio and cost ratio yield higher oscillation in early epochs. Finally, we observe a smooth reduction in training and validation costs for all data sets and experimental settings and Figure 1(b) show the observed decrease on the IDC data set. 

\section{Conclusions}
We proposed a novel cost-sensitive learning method that combines both algorithmic and decision level approaches, aiming to achieve minimum misclassification costs derived from the consequences of various errors. The adaptive cost-sensitive learning (AdaCSL) algorithm was developed to apply the proposed method and 
adaptively adjusts the loss function such that the classifier would be directed to reduce local high misclassification costs in the validation data set. 
Our results on four popular data sets, 
with different setups, show that a CNN classifier with the AdaCSL algorithm outperformed other alternative methods, with an average cost reduction of 20\%.  Furthermore, the proposed algorithm achieved competitive accuracy in experimental setups that reflect an imbalance ratio equal to the cost ratio (i.e., the conventional class-imbalance problems), with better accuracy results in the class-balanced data set.
Future works can apply the proposed cost-sensitive learning method and AdaCSL algorithm to design a variety of cost-sensitive learning models with different loss functions. Another possible research direction is to exploit the method and adjust the AdaCSL algorithm to multi-class classification tasks. 

\vskip -2ex

\FloatBarrier
\newpage
\Urlmuskip=0mu plus 1mu\relax

\bibliography{adacsl_bib}

\begin{thebibliography}{32}
\providecommand{\natexlab}[1]{#1}
\providecommand{\url}[1]{\texttt{#1}}
\expandafter\ifx\csname urlstyle\endcsname\relax
  \providecommand{\doi}[1]{doi: #1}\else
  \providecommand{\doi}{doi: \begingroup \urlstyle{rm}\Url}\fi

\bibitem[Aurelio et~al.(2019)Aurelio, de~Almeida, de~Castro, and
  Braga]{aurelio2019learning}
Aurelio, Y.~S., de~Almeida, G.~M., de~Castro, C.~L., and Braga, A.~P.
\newblock Learning from imbalanced data sets with weighted cross-entropy
  function.
\newblock \emph{Neural Processing Letters}, 50\penalty0 (2):\penalty0
  1937--1949, 2019.

\bibitem[Buda et~al.(2018)Buda, Maki, and Mazurowski]{buda2018systematic}
Buda, M., Maki, A., and Mazurowski, M.~A.
\newblock A systematic study of the class imbalance problem in convolutional
  neural networks.
\newblock \emph{Neural Networks}, 106:\penalty0 249--259, 2018.

\bibitem[Chaabane et~al.(2019)Chaabane, Guermazi, and
  Hammami]{chaabane2019enhancing}
Chaabane, I., Guermazi, R., and Hammami, M.
\newblock Enhancing techniques for learning decision trees from imbalanced
  data.
\newblock \emph{Advances in Data Analysis and Classification}, pages 1--69,
  2019.

\bibitem[Deng et~al.(2009)Deng, Dong, Socher, Li, Li, and
  Fei-Fei]{deng2009imagenet}
Deng, J., Dong, W., Socher, R., Li, L.-J., Li, K., and Fei-Fei, L.
\newblock Imagenet: A large-scale hierarchical image database.
\newblock In \emph{2009 IEEE conference on computer vision and pattern
  recognition}, pages 248--255. Ieee, 2009.

\bibitem[Domingos(1999)]{domingos1999metacost}
Domingos, P.
\newblock Metacost: A general method for making classifiers cost-sensitive.
\newblock In \emph{Proceedings of the fifth ACM SIGKDD international conference
  on Knowledge discovery and data mining}, pages 155--164, 1999.

\bibitem[Dong et~al.(2020)Dong, Zhu, and Zhang]{dong2020cost}
Dong, H., Zhu, B., and Zhang, J.
\newblock A cost-sensitive active learning for imbalance data with uncertainty
  and diversity combination.
\newblock In \emph{Proceedings of the 2020 12th International Conference on
  Machine Learning and Computing}, pages 218--224, 2020.

\bibitem[Elkan(2001)]{elkan2001foundations}
Elkan, C.
\newblock The foundations of cost-sensitive learning.
\newblock In \emph{International joint conference on artificial intelligence},
  volume~17, pages 973--978. Lawrence Erlbaum Associates Ltd, 2001.

\bibitem[Geng and Luo(2018)]{geng2018cost}
Geng, Y. and Luo, X.
\newblock Cost-sensitive convolution based neural networks for imbalanced
  time-series classification.
\newblock \emph{arXiv preprint arXiv:1801.04396}, 2018.

\bibitem[He et~al.(2016)He, Zhang, Ren, and Sun]{he2016deep}
He, K., Zhang, X., Ren, S., and Sun, J.
\newblock Deep residual learning for image recognition.
\newblock In \emph{Proceedings of the IEEE conference on computer vision and
  pattern recognition}, pages 770--778, 2016.

\bibitem[Hern{\'a}ndez-Orallo et~al.(2013)Hern{\'a}ndez-Orallo, Flach, and
  Ferri]{hernandez2013roc}
Hern{\'a}ndez-Orallo, J., Flach, P., and Ferri, C.
\newblock Roc curves in cost space.
\newblock \emph{Machine learning}, 93\penalty0 (1):\penalty0 71--91, 2013.

\bibitem[Janowczyk and Madabhushi(2016)]{janowczyk2016deep}
Janowczyk, A. and Madabhushi, A.
\newblock Deep learning for digital pathology image analysis: A comprehensive
  tutorial with selected use cases.
\newblock \emph{Journal of pathology informatics}, 7, 2016.

\bibitem[Khan et~al.(2017)Khan, Hayat, Bennamoun, Sohel, and
  Togneri]{khan2017cost}
Khan, S.~H., Hayat, M., Bennamoun, M., Sohel, F.~A., and Togneri, R.
\newblock Cost-sensitive learning of deep feature representations from
  imbalanced data.
\newblock \emph{IEEE transactions on neural networks and learning systems},
  29\penalty0 (8):\penalty0 3573--3587, 2017.

\bibitem[Kong et~al.(2020)Kong, Xu, Wang, and Plumbley]{kong2020sound}
Kong, Q., Xu, Y., Wang, W., and Plumbley, M.~D.
\newblock Sound event detection of weakly labelled data with cnn-transformer
  and automatic threshold optimization.
\newblock \emph{IEEE/ACM Transactions on Audio, Speech, and Language
  Processing}, 28:\penalty0 2450--2460, 2020.

\bibitem[Lan et~al.(2020)Lan, Luo, Chai, Chai, Zhang, and Zhang]{lan2020novel}
Lan, M., Luo, J., Chai, S., Chai, R., Zhang, C., and Zhang, B.
\newblock A novel industrial intrusion detection method based on
  threshold-optimized cnn-bilstm-attention using roc curve.
\newblock In \emph{2020 39th Chinese Control Conference (CCC)}, pages
  7384--7389. IEEE, 2020.

\bibitem[Lin et~al.(2017)Lin, Goyal, Girshick, He, and
  Doll{\'a}r]{lin2017focal}
Lin, T.-Y., Goyal, P., Girshick, R., He, K., and Doll{\'a}r, P.
\newblock Focal loss for dense object detection.
\newblock In \emph{Proceedings of the IEEE international conference on computer
  vision}, pages 2980--2988, 2017.

\bibitem[Ling and Sheng(2008)]{ling2008cost}
Ling, C.~X. and Sheng, V.~S.
\newblock Cost-sensitive learning and the class imbalance problem.
\newblock \emph{Encyclopedia of machine learning}, 2011:\penalty0 231--235,
  2008.

\bibitem[Ling et~al.(2004)Ling, Yang, Wang, and Zhang]{ling2004decision}
Ling, C.~X., Yang, Q., Wang, J., and Zhang, S.
\newblock Decision trees with minimal costs.
\newblock In \emph{Proceedings of the twenty-first international conference on
  Machine learning}, page~69, 2004.

\bibitem[Maimon and Rokach(2014)]{maimon2014data}
Maimon, O.~Z. and Rokach, L.
\newblock \emph{Data mining with decision trees: theory and applications},
  volume~81.
\newblock World scientific, 2014.

\bibitem[Masnadi-Shirazi and Vasconcelos(2010)]{masnadi2010cost}
Masnadi-Shirazi, H. and Vasconcelos, N.
\newblock Cost-sensitive boosting.
\newblock \emph{IEEE Transactions on pattern analysis and machine
  intelligence}, 33\penalty0 (2):\penalty0 294--309, 2010.

\bibitem[Mohanty et~al.(2016)Mohanty, Hughes, and
  Salath{\'e}]{mohanty2016using}
Mohanty, S.~P., Hughes, D.~P., and Salath{\'e}, M.
\newblock Using deep learning for image-based plant disease detection.
\newblock \emph{Frontiers in plant science}, 7:\penalty0 1419, 2016.

\bibitem[Pandey and Janghel(2019)]{pandey2019automatic}
Pandey, S.~K. and Janghel, R.~R.
\newblock Automatic detection of arrhythmia from imbalanced ecg database using
  cnn model with smote.
\newblock \emph{Australasian physical \& engineering sciences in medicine},
  42\penalty0 (4):\penalty0 1129--1139, 2019.

\bibitem[Paszke et~al.(2019)Paszke, Gross, Massa, Lerer, Bradbury, Chanan,
  Killeen, Lin, Gimelshein, Antiga, et~al.]{paszke2019pytorch}
Paszke, A., Gross, S., Massa, F., Lerer, A., Bradbury, J., Chanan, G., Killeen,
  T., Lin, Z., Gimelshein, N., Antiga, L., et~al.
\newblock Pytorch: An imperative style, high-performance deep learning library.
\newblock \emph{arXiv preprint arXiv:1912.01703}, 2019.

\bibitem[Raj et~al.(2016)Raj, Magg, and Wermter]{raj2016towards}
Raj, V., Magg, S., and Wermter, S.
\newblock Towards effective classification of imbalanced data with
  convolutional neural networks.
\newblock In \emph{IAPR Workshop on Artificial Neural Networks in Pattern
  Recognition}, pages 150--162. Springer, 2016.

\bibitem[Ribeiro et~al.(2016)Ribeiro, Singh, and Guestrin]{ribeiro2016should}
Ribeiro, M.~T., Singh, S., and Guestrin, C.
\newblock " why should i trust you?" explaining the predictions of any
  classifier.
\newblock In \emph{Proceedings of the 22nd ACM SIGKDD international conference
  on knowledge discovery and data mining}, pages 1135--1144, 2016.

\bibitem[Sze-To and Wong(2017)]{sze2017weight}
Sze-To, A. and Wong, A.~K.
\newblock A weight-selection strategy on training deep neural networks for
  imbalanced classification.
\newblock In \emph{International Conference Image Analysis and Recognition},
  pages 3--10. Springer, 2017.

\bibitem[Ting(2002)]{ting2002instance}
Ting, K.~M.
\newblock An instance-weighting method to induce cost-sensitive trees.
\newblock \emph{IEEE Transactions on Knowledge and Data Engineering},
  14\penalty0 (3):\penalty0 659--665, 2002.

\bibitem[Tschandl et~al.(2018)Tschandl, Rosendahl, and
  Kittler]{tschandl2018ham10000}
Tschandl, P., Rosendahl, C., and Kittler, H.
\newblock The ham10000 dataset, a large collection of multi-source
  dermatoscopic images of common pigmented skin lesions.
\newblock \emph{Scientific data}, 5\penalty0 (1):\penalty0 1--9, 2018.

\bibitem[Wang et~al.(2016)Wang, Liu, Wu, Cao, Meng, and
  Kennedy]{wang2016training}
Wang, S., Liu, W., Wu, J., Cao, L., Meng, Q., and Kennedy, P.~J.
\newblock Training deep neural networks on imbalanced data sets.
\newblock In \emph{2016 international joint conference on neural networks
  (IJCNN)}, pages 4368--4374. IEEE, 2016.

\bibitem[Wu et~al.(2018)Wu, Guo, Lin, Yu, and Ji]{wu2018weighted}
Wu, Z., Guo, Y., Lin, W., Yu, S., and Ji, Y.
\newblock A weighted deep representation learning model for imbalanced fault
  diagnosis in cyber-physical systems.
\newblock \emph{Sensors}, 18\penalty0 (4):\penalty0 1096, 2018.

\bibitem[Yonsei-University(2019)]{RFFD}
Yonsei-University.
\newblock Real and fake face detection, 2019.
\newblock URL
  \url{https://www.kaggle.com/ciplab/real-and-fake-face-detection/}.

\bibitem[Zhang et~al.(2018)Zhang, Tan, Li, and Hong]{zhang2018cost}
Zhang, C., Tan, K.~C., Li, H., and Hong, G.~S.
\newblock A cost-sensitive deep belief network for imbalanced classification.
\newblock \emph{IEEE transactions on neural networks and learning systems},
  30\penalty0 (1):\penalty0 109--122, 2018.

\bibitem[Zhao and Li(2017)]{zhao2017cost}
Zhao, H. and Li, X.
\newblock A cost sensitive decision tree algorithm based on weighted class
  distribution with batch deleting attribute mechanism.
\newblock \emph{Information Sciences}, 378:\penalty0 303--316, 2017.

\end{thebibliography}
\end{document}